%% file: acl_latex.tex
\documentclass[11pt]{article}

\usepackage[final]{acl}

\usepackage{times}
\usepackage{latexsym}
\usepackage[T1]{fontenc}
\usepackage[utf8]{inputenc}
\usepackage{microtype}
\usepackage{inconsolata}
\usepackage{graphicx}

\usepackage{amsmath}
\usepackage{booktabs}
\usepackage{tabularx}
\usepackage{subcaption}

\title{Parameter Golf: What Really Works?}

\author{Prashanna Mani Paudel and Shivanand Venkanna Sheshappanavar\\ Geometric Intelligence Research Lab.,\\ Department of Electrical Engineering and Computer Science\\
University of Wyoming\\
{\tt\small {\{ppaudel,ssheshap}\}@uwyo.edu}}

\begin{document}

\makeatletter
\newsavebox{\arcteaser}
\savebox{\arcteaser}{\includegraphics[width=\textwidth]{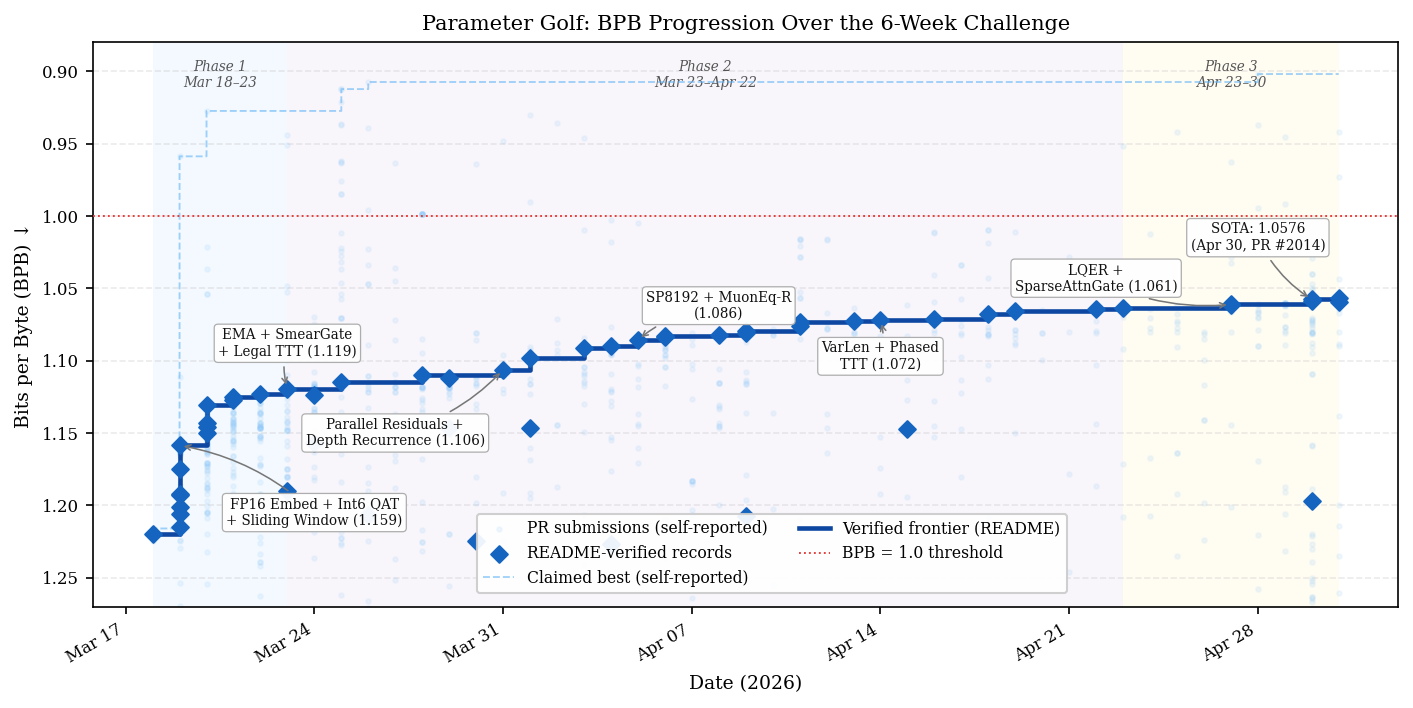}}
\twocolumn[%
  \@maketitle
  \begin{center}
    \usebox{\arcteaser}
    \captionof{figure}{\textbf{Parameter Golf: 43 days, 1,810 submissions, BPB improved from 1.2244 to 1.058 under a fixed artifact-size budget.} Shaded bands (three improvement phases). Labeled points (README-verified milestones).}
    \label{fig:arc}
  \end{center}
]
\setcounter{footnote}{0}

\input{sec/0_abstract}

\input{sec/1_intro}
\input{sec/2_related_work}
\input{sec/3_dataset}
\input{sec/4_challenge}
\input{sec/5_phases}
\input{sec/6_impact}
\input{sec/7_conclusion}

\input{sec/limitations}

\bibliography{custom}

\appendix
\input{sec/appendix}

\end{document}

%% file: sec/0_abstract.tex
\begin{abstract}
How far can a language model improve under a strict artifact budget? \emph{Parameter Golf} posed this question as an open community challenge in which participants trained the best language model, with the complete artifact (training code + compressed weights) required to fit within 16\,MB and to be trained in under ten minutes on $8\times$H100 SXM GPUs. Quality was measured in bits-per-byte (BPB), the average number of bits required to encode each byte of unseen text. We analyze 2,037 pull requests and 1,430 clean-scored submissions from the contest, build a taxonomy of 84 optimization techniques, and measure each technique's contribution to BPB. The verified leaderboard score dropped from 1.2244 to 1.058 BPB across three phases—a 13.6\% reduction, despite individual techniques rarely improving BPB by more than 1\%. We show that most techniques' gains shrink when re-measured among competitive submissions, isolating the few methods that help regardless of the surrounding stack. Code and data are available at \href{https://github.com/PMP56/pmgolf-analysis}{\texttt{github.com/PMP56/pmgolf-analysis}}.
\end{abstract}

%% file: sec/1_intro.tex
\section{Introduction}
\label{sec:intro}

Language model research largely follows a scaling paradigm—more parameters, data, and compute yield better performance. Yet this approach is impractical in settings with strict memory and latency constraints, including on-device inference, privacy-preserving local LMs, and real-time embedded systems operating on kilobyte-to-megabyte budgets. This work explores what is achievable within these extreme limits and how closely current methods approach that frontier. This question is central to BabyLM~\citep{warstadt2023babylm}, which likewise asks how much language-model capability a fixed, severe resource constraint allows; \emph{Parameter Golf} constrains artifact size and training time where BabyLM constrains the training data.

\emph{Parameter Golf} is an open community challenge posted by OpenAI in March 2026~\citep{openai2026parametergolf} that provides a setting to study this problem. Participants compete to minimize bits-per-byte (BPB) on the FineWeb~\citep{penedo2024fineweb} validation dataset, subject to two hard constraints: (i)~the complete submission artifact (training code + compressed model weights) must not exceed 16\,MB, and (ii)~training must complete in under ten minutes on $8\times$H100 SXM GPUs. The metric BPB is tokenizer-agnostic, allowing a fair comparison of submissions with different vocabularies. The challenge drew over 2,000 submissions from the global ML community in six weeks, creating an unprecedented public record of incremental discovery.

\paragraph{Why is this scientifically valuable?} Traditional empirical ML research often takes months or years to reach publication, with papers reporting only final outcomes. Under tight resource constraints, \emph{Parameter Golf} compresses this cycle to days: each submission becomes a natural ablation against the current frontier. The resulting history forms a structured dataset for \emph{retrospective meta-analysis}, enabling the estimation of technique-level BPB effects across thousands of submissions.

%% file: sec/2_related_work.tex
\section{Related Work}
\label{sec:related_work}

\subsection{Constrained LM Challenges}
\label{sec:rw_constrained}

Parameter Golf belongs to a family of community challenges built around the scaling triad $L(N, D, T)$, where $N$ is the model size, $D$ is the dataset size, and $T$ is the training compute. Each challenge fixes one of these axes and asks participants to optimize a model within that constraint. NanoGPT Speedrun~\citep{jordan2024nanogptspeedrun} fixes a target validation loss and minimizes training time; NanoGPT Slowrun~\citep{qlabs2025slowrun} fixes the dataset and minimizes the achievable loss. BabyLM~\citep{warstadt2023babylm} fixes a 10M- or 100M-word corpus and asks participants to train the best model on it. TinyStories~\citep{eldan2023tinystories} pushes the parameter axis to the extreme low end, showing coherent English emerges at sizes far below the frontier. Together, they shift attention from ``how big can we scale?'' to ``how much can we achieve within a fixed constraint?''

\subsection{Transformer Ablation Studies}
\label{sec:rw_ablations}

Architectural improvements are typically evaluated in isolation, leaving open the question of whether they remain effective when combined. \citet{narang2021dotransformermods} is the most direct precedent: they tested dozens of published modifications under matched conditions and found that almost none robustly outperformed a vanilla Transformer. Our within-frontier analysis (Section~\ref{sec:within_frontier}) reaches a similar conclusion from a complementary angle: rather than re-running controlled experiments, we ask which techniques distinguish the strongest submissions from one another. Both studies found that most architectural changes look promising in isolation but not in an already competitive stack. Primer~\citep{so2021primer} uses neural architecture search to find the small set of changes that do survive. The scaling-law literature~\citep{kaplan2020scaling, hoffmann2022chinchilla} supports this view: within a reasonable architectural family, the dominant levers are model size, data, and compute. Parameter Golf is consistent with that: when one axis is locked, what separates the frontier is not architectural novelty but careful allocation of the remaining budget.


\subsection{Hybrid and Non-Parametric LMs}
\label{sec:rw_hybrid}

The within-frontier analysis finds that classical $n$-gram blending is the only technique to retain a clearly positive gain across already-competitive submissions. This result has a long pedigree. Katz backoff~\citep{katz1987backoff} and PPM~\citep{cleary1984ppm} predate neural LMs by decades and remain strong at byte-level prediction. Modern work has revisited the same idea under different framings: kNN-LM~\citep{khandelwal2020knnlm} adds a nearest-neighbor cache over training-set continuations to improve perplexity; RETRO~\citep{borgeaud2022retro} scales the same idea to trillion-token retrieval; and Infini-gram~\citep{liu2024infinigram} shows a large $n$-gram model remains useful even alongside frontier LMs. These results suggest that the neural model and the $n$-gram lookup hold complementary information and that combining them yields performance gains that neither could achieve alone. 

%% file: sec/3_dataset.tex
\section{Dataset and Methodology}
\label{sec:dataset}

\subsection{Data Collection}

We collected all 2,037 pull requests from the \texttt{openai/parameter-golf} repository (March 18--April 30, 2026) via the GitHub REST API. We downloaded each PR's \texttt{README.md} and \texttt{submission.json}, which carry reported scores and architecture metadata. De-duplicating to one row per PR leaves 1,810 rows. For each, we extract a score from five sources in priority order, stopping at the first match (Appendix Figure~\ref{fig:pipeline}): the official leaderboard, the \texttt{val\_bpb} field of \texttt{submission.json}, a README score, PR body text, and finally an LLM reviewer, which adjudicates the 481~PRs the first two sources miss (Section~\ref{sec:detect}). In total, 1,478 submissions carry a valid score; excluding 48 with BPB below 0.9 leaves \textbf{1,430 clean-scored submissions} used in all subsequent analyses.

\begin{table}[t]
  \centering
  \caption{Score-source tiers for the 1,810 submissions.}
  \label{tab:provenance}
  \small
  \begin{tabular}{@{}lrl@{}}
    \toprule
    Tier & Count & Source \\
    \midrule
    Verified leaderboard & 61    & OpenAI-verified \\
    Submission-JSON      & 1,268 & \texttt{val\_bpb} field \\
    README-extracted     & 104   & README headline \\
    PR-body              & 31    & self-reported \\
    LLM-reviewed         & 14    & LLM-recovered \\
    \midrule
    Unscored             & 332   & No score found \\
    \midrule
    \textbf{Total}       & \textbf{1,810} & \\
    \bottomrule
  \end{tabular}
\end{table}

\subsection{Technique Classification}

For each PR, we detect techniques using deterministic keyword matching across three text sources: the PR title, the README body, and the \texttt{submission.json} metadata, which encodes architectural hyperparameters. README mining identified 84 binary technique flags across five categories: Quantization, Architecture, Tokenization, Training, and Evaluation, listed in full in Appendix~\ref{app:flags}.

\subsection{Technique Impact Metric}

For each technique $k$, we compute:
\begin{equation}
  \Delta_k = \overline{\text{BPB}}_{k=0} - \overline{\text{BPB}}_{k=1}
\end{equation}

where $\overline{\text{BPB}}_{k=0}$ and $\overline{\text{BPB}}_{k=1}$ are the mean BPB across clean submissions not using and using technique $k$, respectively. A positive $\Delta_k$ means users of technique $k$ have a lower (better) BPB on average. We compute $\Delta_k$ over the 1,430 clean-scored submissions, excluding the 48 below 0.9~BPB, which are suspected of validation-data leakage and whose abnormally low scores would inflate the apparent gain of any technique they use.

This metric is observational, not causal, and two systematic biases affect it in opposite directions. Techniques introduced late appear only in submissions that already incorporate all prior improvements, inflating their apparent $\Delta_k$. Near-universal techniques such as sliding-window evaluation, Quantization-Aware Training (QAT), and Exponential Moving Average (EMA) yield near-zero $\Delta_k$ due to the lack of a competitive ``without'' group. $\Delta_k$ should therefore be read as a correlation with strong submissions, not a controlled ablation. To separate the two, Section~\ref{sec:impact_summary} recomputes $\Delta_k$ within the competitive frontier alone, and we discuss the confounding explicitly per technique.

\subsection{Accuracy of BPB Score and Technique Detection}
\label{sec:detect}

\paragraph{Score validation:} We validated the extracted BPB score in two passes. A sample of 70 PRs, stratified across sources, was manually reviewed against the parser output. One miss was found, yielding 98.6\% accuracy for \emph{score extraction}; this was later fixed. This accuracy refers only to score parsing; the 84 technique flags are validated separately, as detailed below. The 481~PRs not covered by leaderboards or \texttt{submission.json} are more error-prone, so a language-model reviewer double-checked them. Given the PR's README and \texttt{submission.json} candidates, it assessed whether each candidate's score was genuine. Of the 481, 389 carried a regex-extracted candidate from the README or PR body text: the reviewer confirmed 135, corrected 14 parser errors, and rejected 240 for lacking a valid score. The remaining 92 carried no recoverable score. The 149 retained scores (135 confirmed plus 14 corrected) match the tiers extracted from the README, PR body, and LLM review, as shown in Table~\ref{tab:provenance}.

\paragraph{Technique validation:} For technique presence, all 61 verified leaderboard entries were checked by hand, and the keyword detectors were tightened after auditing their matches on all 1,810 PRs to remove substring false positives. Three residual error sources remain: code comments that name a technique without implementing it, non-standard terminology, and multi-contribution PRs with an ambiguous primary technique. Since we did not hand-label every flag, some noise remains, so we treat each $\Delta_k$ as approximate, relying on the \emph{ranking} of techniques and the within-frontier test (Section~\ref{sec:within_frontier}) rather than small differences in values.

%% file: sec/4_challenge.tex
\section{The Parameter Golf Challenge}
\label{sec:challenge}

\subsection{Evaluation Metric: Bits-per-Byte}

BPB measures how many bits a language model needs to encode each byte of a corpus under an arithmetic coding scheme derived from the model's predicted probability distributions. Formally, for a model assigning token probability $p_t$ at position $t$:

\begin{equation}
\mathrm{BPB}
= \frac{-\sum_{t} \log_2 p_t}{B_{\text{bytes}}}
= \frac{\mathrm{NLL}_{\text{nats}}}{\ln 2 \cdot \mathrm{BPT}}
\end{equation}

where $B_{\text{bytes}}$ is the corpus byte count, $\mathrm{NLL}_{\text{nats}} = \frac{-\sum_t \ln p_t}{T_{\text{tokens}}}$ is the mean per-token negative log-likelihood in nats (the natural-logarithm analog of bits), and $\mathrm{BPT}$ is bytes per token. The BPT factor makes the BPB tokenizer-agnostic. A 1,024-token vocabulary (BPT $\approx 2.4$) and an 8,192-token vocabulary (BPT $\approx 3.7$) are compared on the same scale, since BPB accounts for how efficiently each tokenizer encodes the data, not just how well the model predicts the next token.

\subsection{Baseline Architecture}

The challenge started with a baseline submission model implementing a compact transformer: 9 transformer layers, a SentencePiece tokenizer with a vocabulary size of 1024, int8 post-training quantization, and zlib compression of the weight file. This achieved 1.2244 BPB, defining the starting point from which all improvements are measured.

\subsection{Score Distribution}

\begin{figure}[t]
  \centering
  \includegraphics[width=\linewidth]{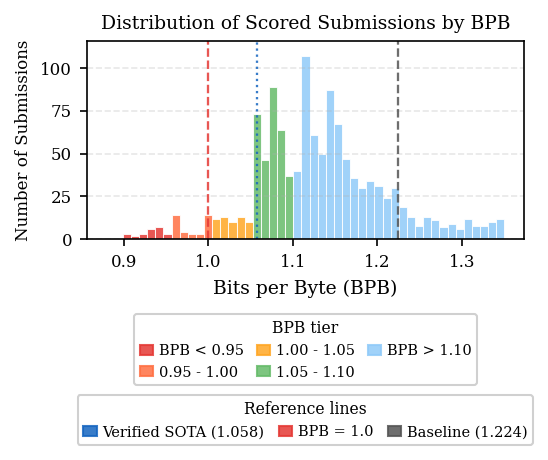}
  \caption{Distribution of BPB scores across the 1,430 clean-scored submissions; 48 leakage-suspected entries below 0.9 BPB are excluded.}
  \label{fig:score_dist}
\end{figure}

Figure~\ref{fig:score_dist} shows how BPB scores are distributed across the 1,430 clean-scored submissions. While Figure~\ref{fig:arc} traces the record-setting frontier, this distribution reveals where the broader community clustered. Most submissions landed in the 1.05--1.20 range, with the competitive frontier (1.05--1.10) containing most verified records. Only a few self-reported submissions crossed the BPB\,=\,1.0 threshold, and none are README-verified.

%% file: sec/5_phases.tex
\section{A 3-Phase Analysis of Parameter Golf}
\label{sec:phases}

We identified three qualitatively distinct improvement phases, each initiated by one or two novel techniques that the community rapidly adopted and built subsequent versions on. We characterize the phases in Table~\ref{tab:phases}, show how each phase's verified improvement splits across technique categories in Figure~\ref{fig:waves}, and describe them below; per-phase technique-impact rankings ($\Delta_k$) are given in Appendix~\ref{app:phase_impact} (Figure~\ref{fig:phase_impact}).

\begin{table*}[t]
  \caption{Three-phase improvement trajectory. BPB decreases steadily across phases (Best result is 1.058 in Phase~3).}
  \label{tab:phases}
  \centering
  \small
  \begin{tabular}{p{2.5cm}llp{8.5cm}}
    \toprule
    \textbf{Phase} & \textbf{Dates} & \textbf{BPB} & \textbf{Dominant techniques} \\
    \midrule
    1. Initial Stack Adoption
      &  3/18--3/23
      & 1.224 $\rightarrow$ 1.12
      & Sliding window, FP16 embed, Int6 QAT-STE, SmearGate, BigramHash, XSA \\
    \addlinespace
    2. Architectural Diversification
      & 3/23--4/22
      & 1.12 $\rightarrow$ 1.064
      & SP8192, depth recurrence, parallel residuals, MuonEq-R, VarLen attention, Legal TTT,
        Gated DeltaNet, Brotli \\
    \addlinespace
    3. Hybrid Emergence
      & 4/23--4/30
      & 1.064 $\rightarrow$ 1.058
      & N-gram backoff, PPM byte mixture, TTT-SLOT, TTT (Global/Phased), LQER,
        sparse attention gate, BOS-boundary fix \\
    \bottomrule
  \end{tabular}
\end{table*}

\begin{figure}[t]
  \centering
  \includegraphics[width=\linewidth]{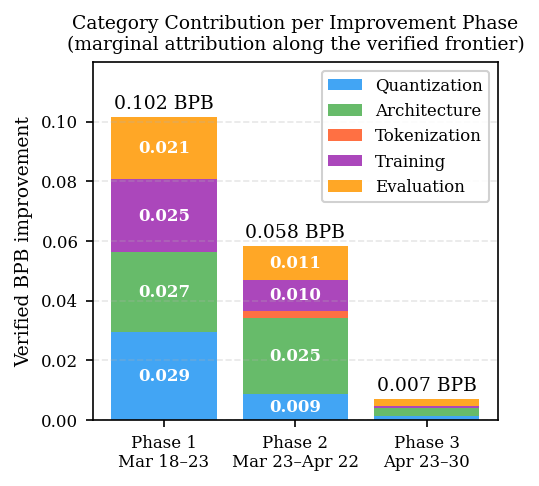}
    \caption{Category contribution per phase. Each record's BPB drop over the previous record is split across its newly added techniques in proportion to their all-data $\Delta_k$ (negatives clamped to zero; drops with no positive-$\Delta_k$ addition split equally). Bar height is the phase's total verified improvement; segments are the per-category shares. The weighting inherits $\Delta_k$'s adoption-timing bias, so the split is indicative rather than exact.}
  \label{fig:waves}
\end{figure}

Appendix Figure~\ref{fig:adoption} tracks technique adoption across 1,810 submissions: quantization reached nearly 90\% by Phase~1, evaluation methods emerged at the Phase~1/2 boundary, and tokenization exceeded 70\% by Phase~3, aligning each category’s rise with its defining phase.

\subsection{Phase 1: Initial Stack Adoption}

The challenge quickly showed that the evaluation strategy can matter as much as model quality. Two major gains in this phase required no architectural or training changes:

\subsubsection{Evaluation strategies}
\textbf{Sliding window evaluation}: Standard chunked evaluation scores non-overlapping 1024-token blocks independently, so early tokens get near-zero context, and the rest average only 512 tokens of context. This inflates BPB by misaligning evaluation with training. A stride-64 sliding window instead gives each scored token 960 tokens of left context without modifying the model. It decreased BPB by $0.032$, the largest improvement with no architectural changes. The cost is $16\times$ more forward passes, making evaluation rival training directly.

\noindent\textbf{FP16 tied embeddings}: The int8 baseline lost precision in the output projection, where small logit errors can substantially increase BPB. Retaining FP16 (16-bit floating point) embeddings preserves accuracy with negligible size overhead. The dataset-wide $\Delta_k = +0.036$ across 131 submissions is small, as FP16 embeddings were largely superseded once later stacks moved to int6 GPTQ~\citep{frantar2023gptq}, a post-training quantizer that rounds each layer's weights to minimize its output error.

\subsubsection{Compression and Quantization}
The first phase of the challenge focused on implementing various quantization strategies to free up space and add more layers to the model, thereby improving its response quality. The implementations that proved to be the most valuable are:

\noindent\textbf{Int6 Quantization-Aware Training with Straight-Through Estimator}: Switching from int8 to int6 reduces storage to 0.75 bytes per weight, freeing roughly 4\,MB that funded MLP expansion from $2\times\to3\times$ and two extra transformer layers (9$\to$11). Since naive int6 post-training degrades accuracy, Quantization-Aware Training (QAT)~\citep{jacob2018qat} simulates quantization in the forward pass, while the Straight-Through Estimator (STE)~\citep{bengio2013ste} passes gradients through the rounding step unchanged. Int6 reached 968 submissions at $\Delta_k = +0.130$, with QAT-STE in 575, carrying the verified leaderboard to approximately 1.146 BPB.

\noindent\textbf{SmearGate}: In capacity-constrained models, high-variance activations in the residual stream propagate through the MLP block and waste computation on uncertain directions. SmearGate adds a learnable soft gate that down-weights these high-variance directions before they reach the MLP, reducing misallocated computation without removing any model components. It reached 548 submissions at a dataset-wide $\Delta_k = +0.105$.

\noindent\textbf{BigramHash Embeddings}: Standard embeddings give the model no signal about local token co-occurrence. BigramHash embeddings augment each token embedding with a learned bigram hash table, adding a cheap inductive bias for short-range dependencies without extra transformer layers. BigramHash reached 635 submissions at $\Delta_k = +0.081$, with negligible compute overhead since the hash table is a simple embedding lookup.


\subsubsection{Attention and Averaging}
Following the compression-driven gains of the first phase, three independently mergeable contributions arrived in rapid succession between March 20 and 23, completing the canonical Phase~1 stack:

\noindent\textbf{XSA (Exclusive Self-Attention)}: XSA replaces full attention in the deepest transformer blocks with a parameter-reduced exclusive-attention variant, freeing bytes that can be reallocated to MLP width or additional layers. This makes XSA a capacity reallocation technique rather than a pure architectural improvement. XSA reached 571 submissions with a dataset-wide $\Delta_k = +0.119$, among the highest of any single technique in this phase.

\noindent\textbf{EMA Weight Averaging}: Stochastic gradient descent produces a noisy parameter trajectory whose final checkpoint is rarely the best generalizing point. Exponential moving average (EMA) weight averaging keeps a running average of checkpoints, producing a smoother estimate that generalizes better, in the spirit of stochastic weight averaging~\citep{izmailov2018swa}. It is applied post-hoc and requires no retraining. EMA reached 767 submissions at $\Delta_k = +0.123$, reflecting consistent improvement and easy integration into existing stacks.

\noindent\textbf{Legal Score-First Test-Time Training}:
Test-time training (TTT)~\citep{sun2020ttt} updates model weights directly on the evaluation corpus before scoring, adapting the model to the target distribution at inference. This variant constrained updates to techniques permitted under the challenge rules, making it the first validated instance of TTT in the competition. Gains were modest initially, but adoption accumulated through Phase~2 to 413 submissions at $\Delta_k = +0.143$, foreshadowing the unrestricted adaptation that would dominate Phase~3.

Phase~1 closed at approximately 1.1194 BPB (March~23) with the community converging on a canonical stack: Int6 QAT-STE, sliding-window eval, XSA, EMA, and SmearGate. Nearly all later submissions treated this stack as the new baseline.

\subsection{Phase 2: Architectural Diversification}

Phase~2 spanned 30 days and improved BPB by approximately 0.06, against 0.10 in the five days of Phase~1. The slowdown is understandable: Phase~1 gains came from fixing evaluation errors and byte inefficiencies, both of which yield substantial improvements at a low implementation cost. Phase~2 required real architectural and training innovations that are harder to find and implement for smaller marginal gains against an already strong base.

\subsubsection{Vocabulary Expansion and Training}
\noindent\textbf{SP8192 Vocabulary}: The baseline tokenizer used a vocabulary of 1,024 tokens, covering approximately 2.4 bytes per token on the evaluation corpus. Expanding to an 8,192-token SentencePiece~\citep{kudo2018sentencepiece} vocabulary raises this to approximately 3.7 bytes per token, a 53\% increase. Since BPB scales inversely with bytes per token at a fixed per-token NLL, the longer tokens would lower BPB by 35\% if the model predicted them as well as before. In practice each token carries more information, so per-token NLL rises and the realized gain is smaller, and the eightfold larger embedding table costs bytes within the 16\,MB budget. SP8192 reached 351 submissions at $\Delta_k = +0.115$, among the largest Phase~2 gains needing no architectural change.

\noindent\textbf{Depth Recurrence}: Depth recurrence loops over selected existing layers multiple times, creating virtual depth from a fixed parameter budget at zero byte cost. The tradeoff is one full forward pass per extra loop. Depth Recurrence reached 405 submissions at $\Delta_k = +0.079$. The relatively modest gain reflects that recurrence introduces optimization difficulties alongside the capacity benefit.

\noindent\textbf{Parallel Residuals (GPT-J Style)}: In standard transformer blocks, the attention sublayer modifies the residual stream before the MLP sees it. Parallel residuals run both sublayers on the same unmodified input, a variant popularized at scale by PaLM~\citep{chowdhery2023palm}, improving gradient flow at the same parameter count. Parallel residuals reached 237 submissions with a dataset-wide $\Delta_k = +0.136$, suggesting that sequential coupling between attention and the MLP is a meaningful bottleneck at this scale.

\noindent\textbf{MuonEq-R}: MuonEq-R is an orthogonalized variant of the Muon optimizer~\citep{jordan2024muon} with equalized update norms, which enforces more uniform parameter utilization. MuonEq-R reached 146 submissions at $\Delta_k = +0.127$, suggesting that the choice of optimizer is a larger lever than commonly assumed at this scale.

\subsubsection{Non-Transformer Architectures and Compression Implementations}

\noindent\textbf{Variable-Length Attention (VarLen)}: VarLen processes variable-length document boundaries within a single forward pass, eliminating the padding waste that fixed-length attention incurs when documents are shorter than the context window. VarLen reached 34 submissions with a dataset-wide $\Delta_k = +0.150$. The low submission count relative to the high $\Delta_k$ suggests that VarLen was difficult to implement correctly, limiting adoption despite a strong signal.

\noindent\textbf{Gated DeltaNet (GDN)}: Gated DeltaNet~\citep{yang2024gateddeltanet} replaces attention entirely with a gated delta-rule recurrence that maintains a fixed-size hidden state, so the full artifact budget goes to model weights rather than a growing key-value cache. GDN reached 50 submissions at a dataset-wide $\Delta_k = +0.076$,  with a best reported (unverified) score near 0.998, showing that non-transformer architectures can be competitive at this scale.

\noindent\textbf{Brotli Weight Compression}: Switching from zlib or zstd-22 compression to Brotli~\citep{alakuijala2018brotli} achieves better compression on neural weight distributions, freeing extra bytes within the 16\,MB constraint with no model changes. Brotli reached 310 submissions with a dataset-wide $\Delta_k = +0.149$. Its high adoption reflects that the switch is a single-line change in the artifact pipeline.

\noindent\textbf{CaseOps/Casefold Tokenizer}: BPE tokenizers treat case variants as distinct tokens. Applying Unicode case-folding before BPE segmentation merges these otherwise-separate tokens, reducing vocabulary sparsity without increasing vocabulary size. CaseOps reached 135 submissions with a dataset-wide $\Delta_k = +0.135$, among the highest in the dataset. This is unusually large for a preprocessing step, suggesting the baseline tokenizer handled the corpus's case-variant sparsity poorly.

By late April, Phase~2 reached approximately 1.064 BPB. Further gains now require combining multiple complementary techniques rather than any single fix, setting the stage for the hybrid, carefully coordinated stacks of Phase~3.

\subsection{Phase 3: Hybrid Emergence}

Phase~3 was qualitatively different from Phase~2. Rather than further refining the neural stack, the top submissions increasingly combined the neural language model with classical $n$-gram sequence prediction and specialized attention mechanisms that directly target the evaluation setting. In eight days, Phase~3 produced a 0.007 BPB improvement in the verified frontier, concentrated in a small number of highly-engineered record-setting stacks rather than broad community adoption.

  \textbf{N-gram Blending}: Phase~3's defining move was to mix neural LM predictions with a classical $n$-gram model at the byte level via a learned interpolation coefficient: $P_\text{mix} = \lambda P_\text{neural} + (1-\lambda) P_\text{ngram}$. Two forms of n-gram blending appeared: an $n$-gram \emph{cache} over recently-seen contexts and an $n$-gram \emph{backoff}~\citep{katz1987backoff} that falls back to shorter contexts, in the manner of PPM~\citep{cleary1984ppm}. Both add zero learnable parameters and store compactly as frequency tables. N-gram backoff reached 28 submissions at $\Delta_k = +0.180$, the highest among all techniques used. The cache form reached 218 submissions at $\Delta_k = +0.147$, and the PPM byte-mixture reached 57 at $\Delta_k = +0.111$. 

\noindent\textbf{Sparse Attention Gate}: The Sparse Attention Gate applies a learned sparsity mask to the attention score matrix before softmax, zeroing low-magnitude entries and concentrating capacity on high-salience pairs. It reached 71 submissions with a dataset-wide $\Delta_k = +0.159$, the second-highest in the dataset. Its late introduction means the value reflects a correlation with a strong late-stage base rather than an isolated contribution. All four lowest verified BPB scores include this technique.

\noindent\textbf{TTT-SLOT (Sequence-Local Online Training)}: Where standard TTT adapts weights to the whole evaluation corpus before scoring, TTT-SLOT applies local weight updates within each scored sequence chunk, adapting to each document before predicting. It reached 242 submissions with a dataset-wide $\Delta_k = +0.139$, the highest adoption of any Phase~3 technique.

\noindent\textbf{TTT (Global/Phased)}: Global/Phased TTT adapts the model to the whole evaluation corpus before scoring, rather than per chunk. The phased variant uses a staged schedule: an initial short-context pass, then passes that extend context and reduce the learning rate. It reached 101 submissions at dataset-wide $\Delta_k = +0.146$. The final SOTA scores use phased TTT with sequences up to 2,816 tokens, driving the verified frontier from 1.061 to 1.058.

\noindent\textbf{LQER (Learned Quantization Error Reduction)}: LQER~\citep{zhang2024lqer} applies a learned post-training correction to the quantization error left by GPTQ~\citep{frantar2023gptq}, recovering precision loss without retraining. It is composable with any quantized model, including the Int6 QAT-STE stacks from Phases~1 and~2. LQER reached 96 submissions with a dataset-wide $\Delta_k = +0.142$.

\noindent\textbf{BOS-Boundary Fix}: An off-by-one error in how beginning-of-sequence (BOS) boundaries were handled during evaluation introduced a systematic scoring bias for documents starting near chunk boundaries. The fix required no model changes, only a correction to the evaluation loop, showing that measurement errors were still silently inflating BPB even in the final phase.

The verified frontier reached 1.058 BPB, the lowest within-deadline score; a later 1.057 entry was submitted on May~1 and accepted by OpenAI.

%% file: sec/6_impact.tex
\section{Technique Impact Summary}
\label{sec:impact_summary}

First, the largest gains come from techniques that step outside the neural model. N-gram backoff leads the dataset at $+0.180$, with the cache form at $+0.147$, and these classical sequence-prediction methods sit alongside the Sparse Attention Gate and the evaluation-time TTT variants at the top of Table~\ref{tab:impact_top}. Classical $n$-gram mixing and direct adaptation to the evaluation distribution outrank any single architectural change.

Second, several widely adopted techniques still post high $\Delta\text{BPB}$. GPTQ and Brotli both compress the weight file to free byte budget, and CaseOps tokenization reaches $+0.135$ across 135 submissions. High values at high adoption reflect that these became part of the canonical stack, so the small ``without'' group is dominated by weaker early submissions.

Third, there is a clear precision sweet spot at int6. Int8-only quantization carries the most negative $\Delta\text{BPB}$ in the dataset, largely from early submissions, but mechanically it spends byte budget that int6 would turn into capacity. Low-bit formats hurt at the other end: Binary (1-bit) at $-0.071$ and Ternary at $-0.013$. Pushing precision either way gives up more than the artifact budget can recover.


\begin{table}[t]
  \caption{Selected techniques from the top and bottom of the all-data $\Delta_k$ ranking across the 1,430 clean-scored submissions; the full top 22 appears in Appendix Figure~\ref{fig:impact}. A positive $\Delta$BPB = lower BPB (better). Techniques not described in Section~\ref{sec:phases} are defined in Appendix~\ref{app:flags}.}
  \label{tab:impact_top}
  \centering
  \small
  \begin{tabular}{llrr}
    \toprule
    \textbf{Technique} & \textbf{Cat.} & \textbf{Count} & $\boldsymbol{\Delta\text{BPB}}$ \\
    \midrule
    N-gram Backoff                  & Eval  & 28  & $+0.180$ \\
    Sparse Attention Gate           & Arch  & 71  & $+0.159$ \\
    GPTQ                            & Quant & 664 & $+0.155$ \\
    AsymLogit                       & Arch  & 28  & $+0.155$ \\
    Attention Output Gate           & Arch  & 38  & $+0.151$ \\
    Variable-Length Attention       & Arch  & 34  & $+0.150$ \\
    Brotli                          & Quant & 310 & $+0.149$ \\
    N-gram Cache                    & Eval  & 218 & $+0.147$ \\
    TTT (Global/Phased)             & Eval  & 101 & $+0.146$ \\
    LQER                            & Quant & 96  & $+0.142$ \\
    \midrule
    Ternary Quantization            & Quant & 188 & $-0.013$ \\
    Universal Transformer           & Arch  & 38  & $-0.098$ \\
    State-Space Model               & Arch  & 60  & $-0.156$ \\
    Int8-only Quantization          & Quant & 258 & $-0.200$ \\
    \bottomrule
  \end{tabular}
\end{table}



\paragraph{\textbf{Within-Frontier Impact:}}
\label{sec:within_frontier}
The $\Delta_k$ values above carry the adoption-timing bias of Section~\ref{sec:dataset}: a technique that became standard in strong stacks is measured against a ``without'' group made largely of weak submissions. To separate timing from contribution, we recompute $\Delta_k$ within the competitive frontier alone, restricting both groups to the 429 submissions scoring below 1.10 BPB. A positive value means the technique separates the best submissions from one another rather than the best era from the worst. Because submission counts span 28 to 968, we attach to each within-frontier $\Delta_k$ a 95\% bootstrap confidence interval (CI) and a permutation $p$-value (10{,}000 resamples), with Benjamini--Hochberg correction~\citep{benjamini1995fdr} across the 70 techniques having at least five frontier submissions.

\begin{figure}[t]
  \centering
  \includegraphics[width=\linewidth]{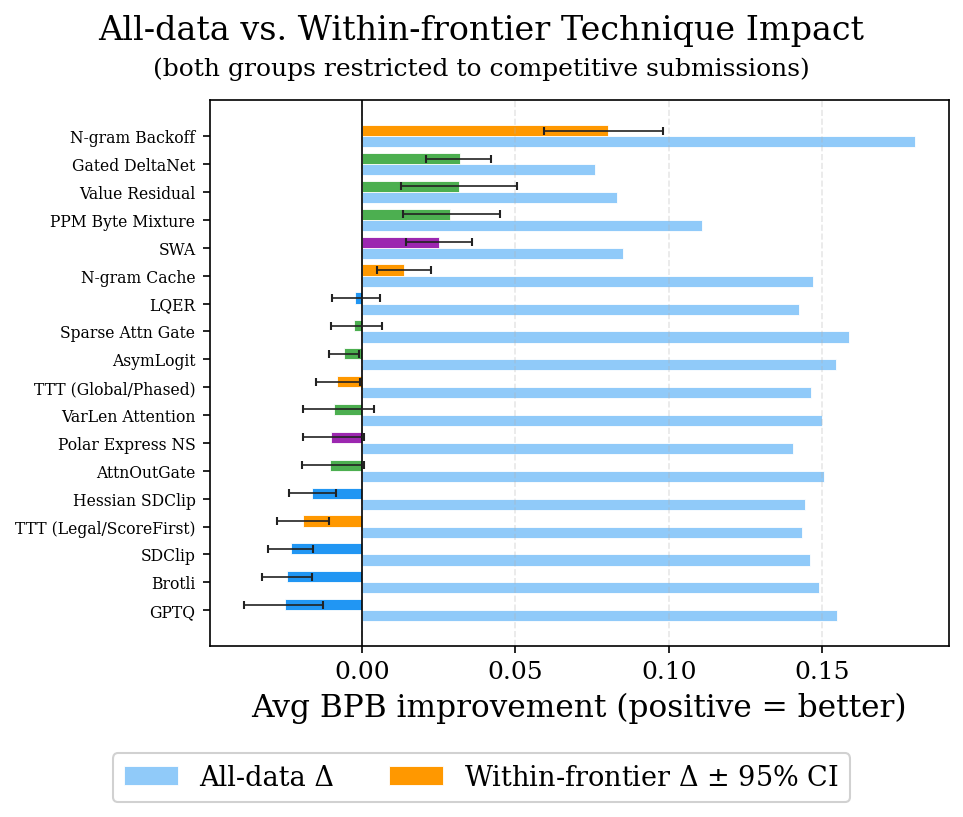}
  \caption{All-data versus within-frontier $\Delta_k$, ordered by within-frontier value. The within-frontier bars carry 95\% bootstrap confidence intervals.}
  \label{fig:within_frontier}
\end{figure}

Figure~\ref{fig:within_frontier} contrasts the two measurements. The effect is striking: almost every technique's apparent gain collapses toward zero or turns slightly negative within the frontier. GPTQ, Brotli, depth recurrence, and the two most common TTT variants, which are all near-universal among strong submissions, fall to small but significantly negative values, confirming that their large all-data $\Delta_k$ reflects era correlation rather than an independent contribution. N-gram backoff is the clear exception, retaining $\Delta_k = +0.080$ (95\% CI $[0.06, 0.10]$, $p < 0.0001$) on an interval lying entirely above every other technique's. This indicates that classical $n$-gram blending improved even submissions that were already competitive. A second tier of techniques (PPM byte mixture, Gated DeltaNet, value residuals, and SWA) remains significantly positive with $\Delta_k$ ranging from $+0.025$ to $+0.032$. These techniques genuinely separate record-setting stacks from the merely strong field.

\begin{figure}[t]
  \centering
  \includegraphics[width=\linewidth]{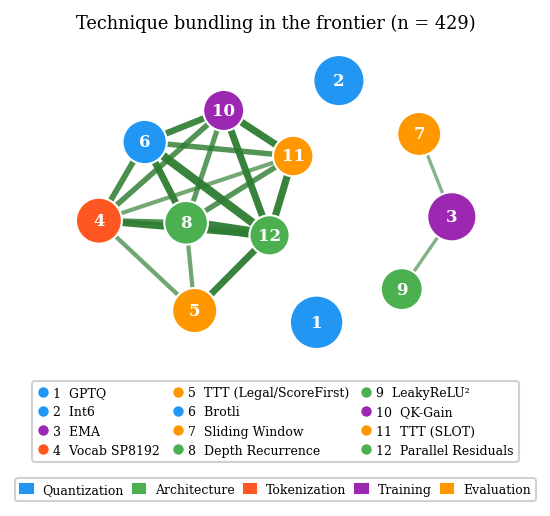}
  \caption{Technique bundling among the 429 frontier submissions (BPB $<$ 1.10). The twelve most prevalent techniques are placed so that pairs with high lift sit close together; circle area is frontier prevalence, color is category, and the number is prevalence rank (1 = most prevalent). An edge is drawn wherever lift, the co-occurrence rate divided by the rate expected under independent adoption, is at least 1.2, with width growing with lift. Exact pairwise values are given in Appendix Figure~\ref{fig:cooccurrence_heatmap}.}
  \label{fig:bundling}
\end{figure}

\paragraph{\textbf{Technique Bundling in the Frontier:}}
Figure~\ref{fig:bundling} maps the twelve most prevalent techniques across the 429 frontier submissions by lift, the rate at which two techniques co-occur relative to that expected under independent adoption. Two groups emerge. The Phase~1 core (GPTQ, int6, EMA, and sliding window; nodes 1--3 and 7) is near-universal, each present in 54--81\% of frontier stacks and all of GPTQ, int6, and EMA in 51\%, yet it forms no bundle: GPTQ, the most prevalent technique, clears the 1.2 edge threshold with nothing; its closest pairing, with Brotli, falls just short. Its members co-occur simply because everyone uses them. The Phase~2--3 additions (nodes 4--6, 8, and 10--12) are the genuine bundle, adopted together well above chance, led by depth recurrence with parallel residuals (lift 1.80) and Brotli with parallel residuals (1.60). Even so, there is no fixed recipe: these twelve techniques form 222 distinct stacks across the 429 submissions, the commonest shared by only 6\%, and only 22\% use all six most-prevalent techniques. This variation gives the within-frontier test its power. Attribution is hardest for the universal core, whose members are too prevalent to leave a comparable ``without'' group, and most tractable for the bundle and periphery, which vary across stacks; the strongest independent contributor, $n$-gram backoff, appears in just 5\% of the frontier.

%% file: sec/7_conclusion.tex
\section{Conclusion}
\label{sec:conclusion}

We analyzed six weeks of \emph{Parameter Golf} across 2,037 pull requests and 1,430 clean-scored submissions. The community lowered the best verified BPB from the naive baseline of 1.2244 to 1.058, a 13.6\% improvement, across three phases: evaluation fixes and quantization, architectural and vocabulary exploration, and finally $n$-gram blending and test-time training. Much of the early improvement came from correcting a deliberately naive baseline by using sliding-window evaluation, FP16 embeddings, and improved tokenization. The headline gain, therefore, reflects cumulative community engineering rather than a single breakthrough.

The throughline is that the frontier moved less through new architectures than by stepping outside the neural model, via $n$-gram blending and test-time adaptation. Int6 quantization made the best use of the byte budget, since it neither wastes bits like int8 nor sacrifices accuracy as lower-bit formats do. And most reported gains shrink once re-measured among competitive submissions, leaving only a few techniques with an independent effect. More broadly, open, continuously scored competitions can function as living ablation studies, revealing which innovations truly advance the frontier and which merely benefit from timing.

%% file: sec/limitations.tex
\section*{Limitations}

Our impact metric $\Delta_k$ is observational, not causal. It measures how strongly a technique co-occurs with low BPB and is biased by adoption timing, which is why we report the within-frontier recomputation alongside it. Even within the frontier, $\Delta_k$ is a marginal association: submissions change several techniques at once and inherit earlier stacks, so it cannot isolate a technique from the bundle it travels with. The bundling analysis in Section~\ref{sec:impact_summary} shows where this bites: the Phase~1 core is so prevalent that its members lack a comparable ``without'' group, so their individual effects are only weakly identified here, whereas the periphery varies across 222 distinct stacks, which is what gives the within-frontier comparison its power. Only a controlled ablation, re-running submissions with one technique toggled, would yield causal effects. Technique detection relies on keyword matching, so it can miss or mislabel techniques that are named only in linked papers, described in non-standard terminology, or mentioned in code comments without implementation. The 61 verified entries were checked by hand, and the detectors were audited on all 1,810 PRs (Section~\ref{sec:detect}). Scores below the verified-leaderboard tier are author-reported and only partially rerun, although those recovered from the README or PR text were additionally checked by a language-model reviewer. We also exclude submissions below 0.9 BPB as suspected validation-data leakage. These caveats limited the precision of individual $\Delta_k$ values but did not affect the qualitative ordering of the strongest and weakest techniques. Finally, all results come from a single challenge setting, so the optimal configurations are tied to its budget and would shift under tighter or looser constraints.

%% file: sec/appendix.tex
\appendix

\section{Appendix}

\subsection{Data Collection Pipeline}
\label{app:pipeline}

\begin{figure*}[t]
  \centering
  \includegraphics[width=0.9\linewidth]{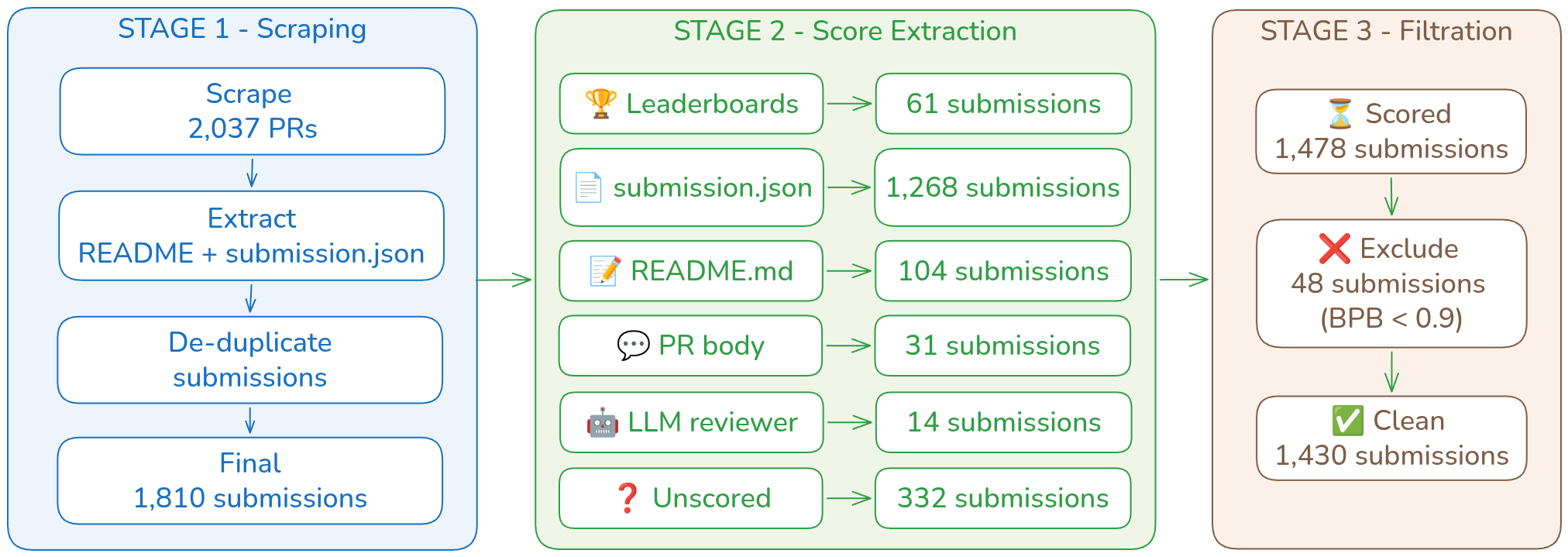}
  \caption{Data collection and score-extraction pipeline. Scores are assigned by trying five sources in order of priority: leaderboard, \texttt{submission.json}, README, PR body, and then an LLM reviewer as a last resort. The leakage filter removes 48 implausible entries (BPB\,$<$\,0.9), leaving \textbf{1,430 clean scored} submissions.}
  \label{fig:pipeline}
\end{figure*}

Figure~\ref{fig:pipeline} details the score-extraction pipeline summarized in Section~\ref{sec:dataset}: each of the 1,810 rows is scored by trying five sources in priority order, after which the leakage filter yields the 1,430 clean scored submissions used throughout.

\subsection{Technique Impact Rankings}
\label{app:phase_impact}

\begin{figure}[t]
  \centering
  \includegraphics[width=\linewidth]{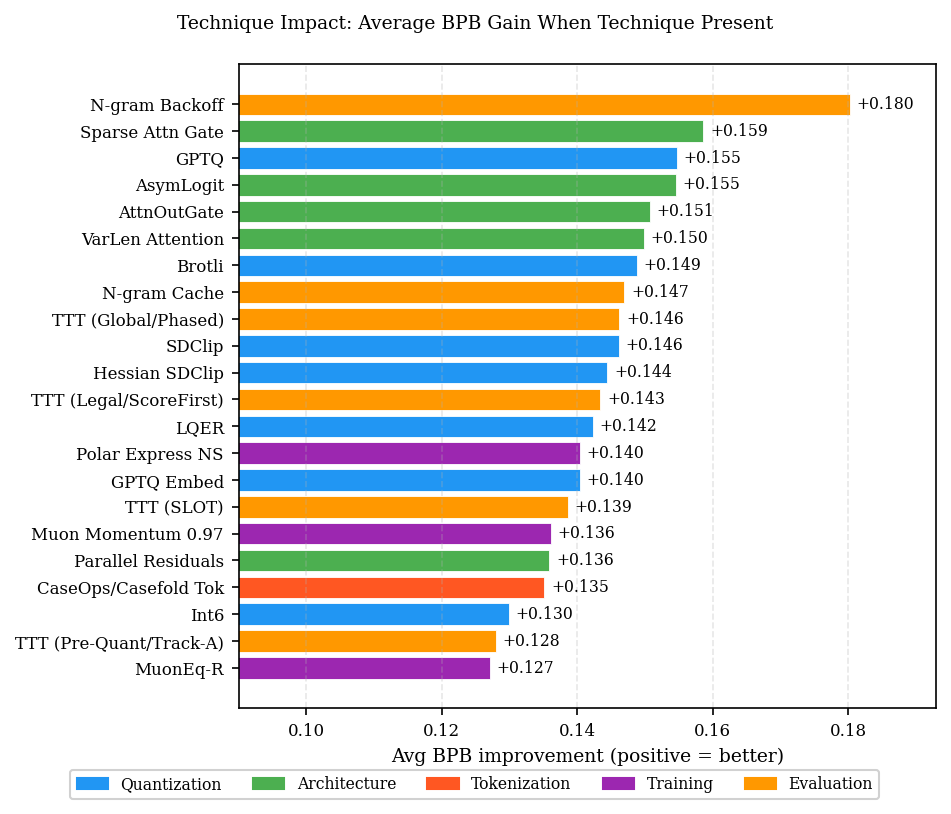}
  \caption{Top-22 techniques ranked by all-data $\Delta_k$ across the 1,430 clean scored submissions. $\Delta_k$ is observational; see Section~\ref{sec:dataset}.}
  \label{fig:impact}
\end{figure}

\begin{figure}[t]
  \centering
  \includegraphics[width=\linewidth]{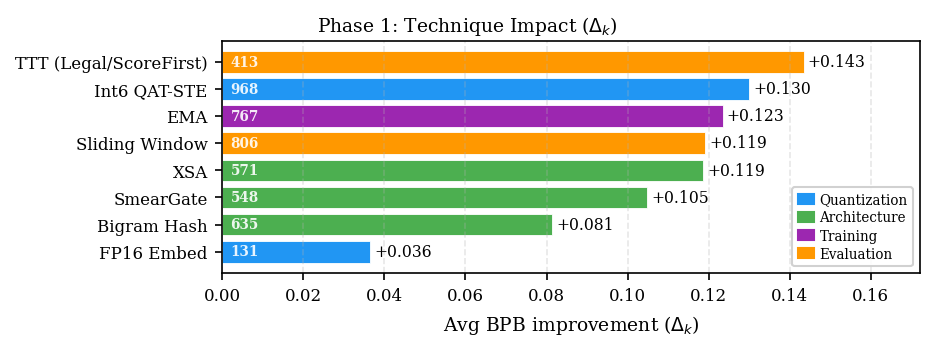}\\[3pt]
  \includegraphics[width=\linewidth]{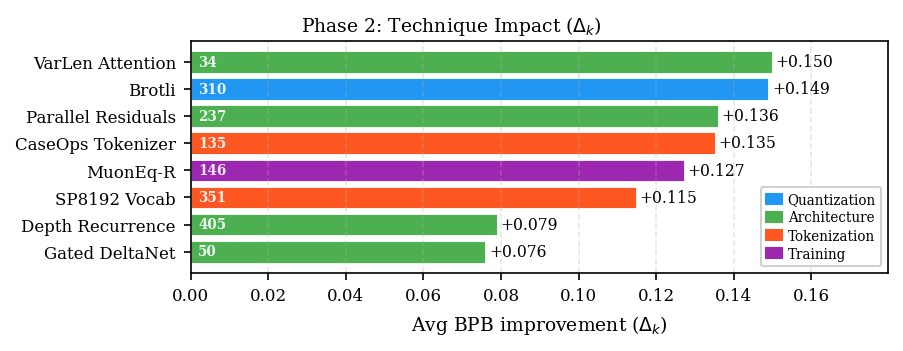}\\[3pt]
  \includegraphics[width=\linewidth]{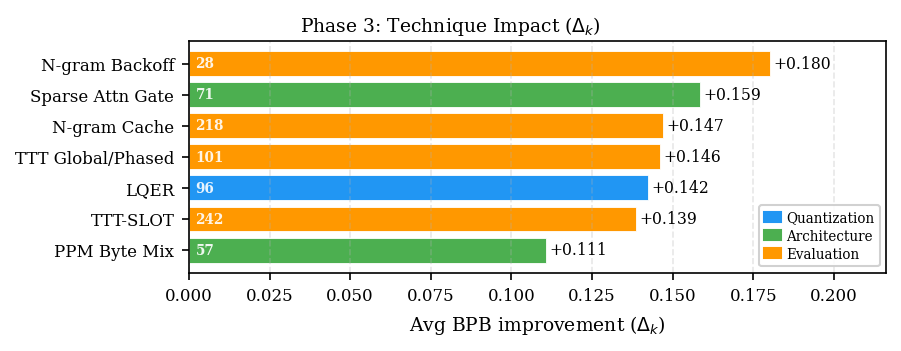}
  \caption{Per-phase technique-impact rankings ($\Delta_k$) across all scored submissions, broken out by phase (top to bottom: Phases 1--3). White bar counts show the number of submissions using each technique.}
  \label{fig:phase_impact}
\end{figure}

Figure~\ref{fig:impact} shows the full top-22 all-data ranking that Table~\ref{tab:impact_top} summarizes, and Figure~\ref{fig:phase_impact} breaks the ranking down by phase, with submission counts shown in white on each bar. They expand the per-technique discussion in Section~\ref{sec:phases}. Like all dataset-wide $\Delta_k$ values, these rankings carry the adoption-timing bias discussed in Section~\ref{sec:dataset}; the within-frontier recomputation in Section~\ref{sec:within_frontier} isolates the genuine contributors.

\subsection{Adoption Dynamics and Category Deep Dives}
\label{app:deepdives}

\begin{figure*}[t]
  \centering
  \includegraphics[width=0.85\linewidth]{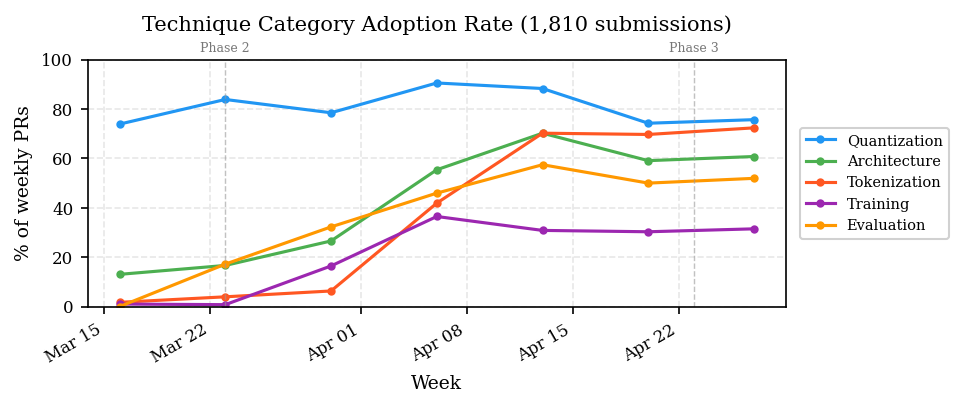}
  \caption{Weekly adoption rate per technique category. Each point is the fraction of that week's PRs using at least one signature technique from the category (counting all 84 flags saturates, since nearly every PR uses one technique per category). Dashed lines mark phase transitions.}
  \label{fig:adoption}
\end{figure*}

\begin{figure*}[t]
  \centering
  \includegraphics[width=\linewidth]{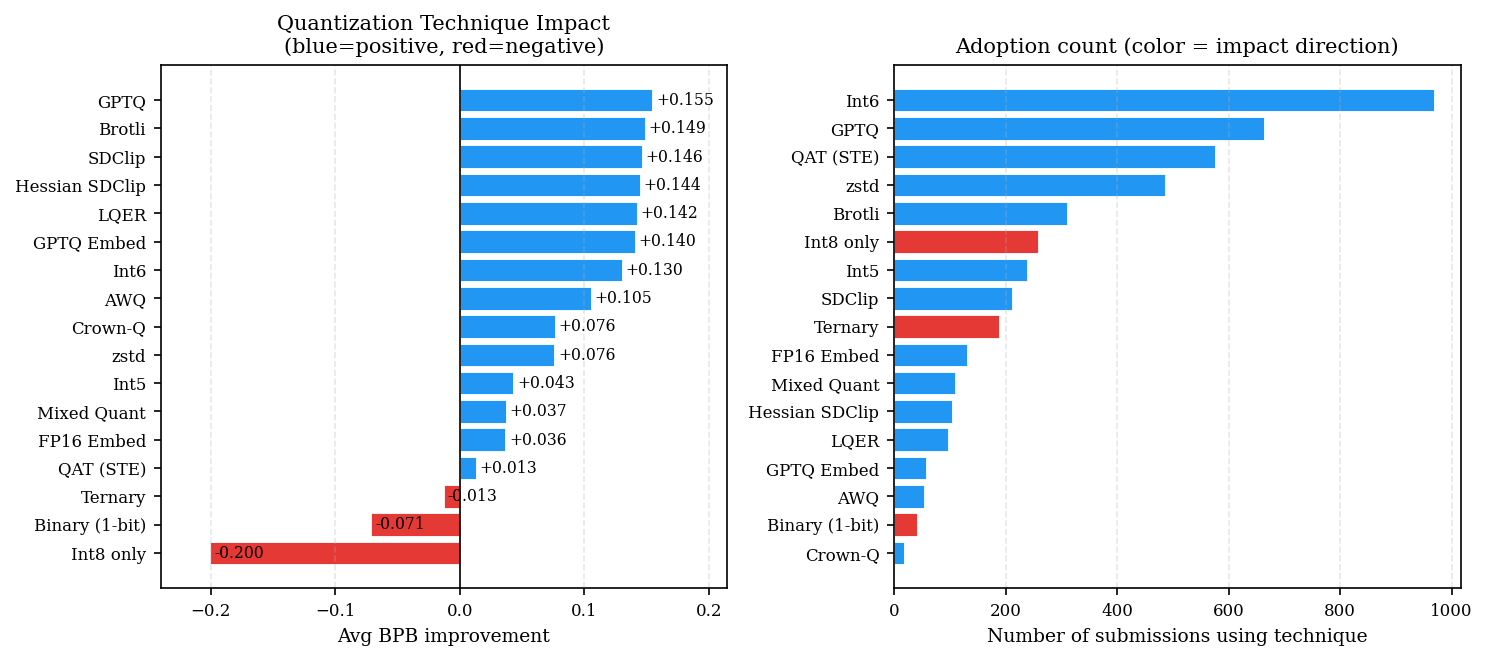}
  \caption{Quantization technique deep dive. \textit{Left}: average BPB improvement $\Delta_k$ for all 17 quantization flags; blue bars are beneficial, red are harmful. \textit{Right}: adoption counts colored by impact direction. Int8-only and Binary (1-bit) correlate with worse BPB despite wide adoption, while GPTQ, Brotli, and SDClip show the strongest gains.}
  \label{fig:quant_deep}
\end{figure*}

\begin{figure}[t]
  \centering
  \includegraphics[width=\linewidth]{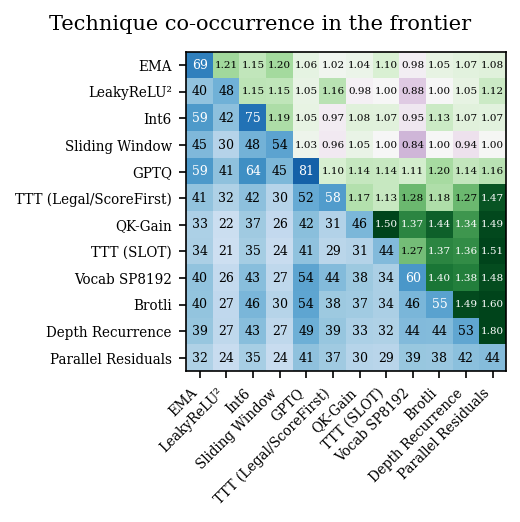}
  \caption{Pairwise co-occurrence of the twelve most prevalent frontier techniques, cluster-ordered so that bundled techniques are adjacent. Lower triangle: percentage of the 429 frontier submissions using both techniques; diagonal: frontier prevalence. Upper triangle: lift (green, $>1$: adopted together more than chance; purple, $<1$: less). This is the exact-number companion to Figure~\ref{fig:bundling}.}
  \label{fig:cooccurrence_heatmap}
\end{figure}

\begin{figure}[h]
  \centering
  \includegraphics[width=\linewidth]{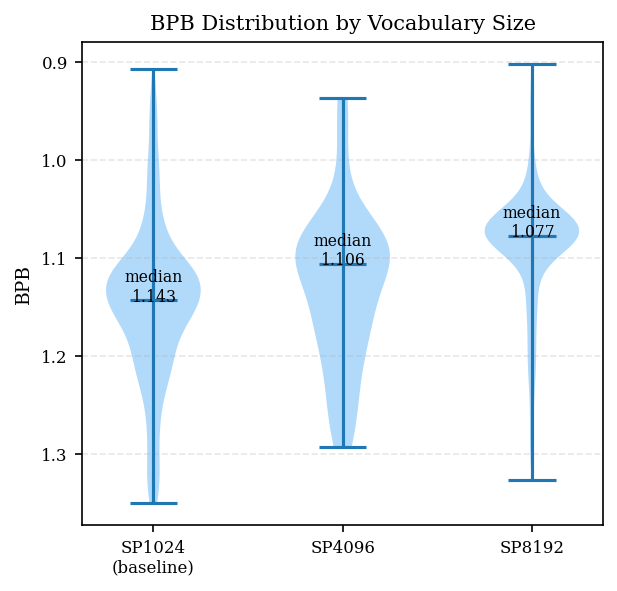}
  \caption{BPB distributions for the three SentencePiece vocabulary sizes; larger vocabularies shift the median downward (SP8192 median 1.077 vs.\ SP1024 baseline 1.143).}
  \label{fig:vocab_size}
\end{figure}

\begin{figure}[h]
  \centering
  \includegraphics[width=\linewidth]{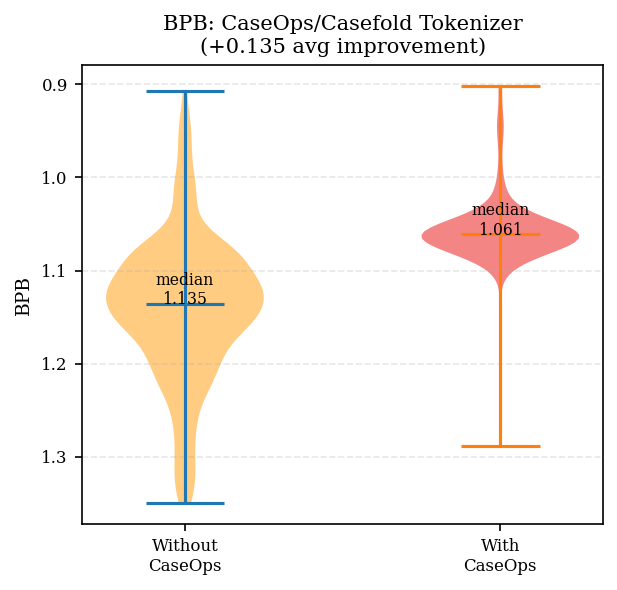}
  \caption{CaseOps/Casefold tokenizer impact: it tightens the upper tail and shifts the median from 1.135 to 1.061, an average $\Delta_k = +0.135$.}
  \label{fig:caseops}
\end{figure}

Figure~\ref{fig:adoption} (discussed in Section~\ref{sec:phases}) traces how quickly each category's signature techniques propagated through the community. Figure~\ref{fig:quant_deep} drills into quantization, and Figures~\ref{fig:vocab_size} and~\ref{fig:caseops} into tokenization, the two categories with the most internal variation in $\Delta_k$. Figure~\ref{fig:cooccurrence_heatmap} gives the exact pairwise co-occurrence and lift values summarized by the bundling map in Figure~\ref{fig:bundling}.

\subsection{Full Technique Flag List}
\label{app:flags}

The 84 binary technique flags are organized into five categories.

\begin{enumerate}
  \item \textbf{Quantization} (17 flags): the precision of weight representation and the compression format of the weight file. Covers int8, int6, int5, FP16 embeddings, QAT-STE, GPTQ~\citep{frantar2023gptq}, GPTQ Embeddings, Ternary, Binary, Mixed-precision, zstd, Brotli, SDClip, Hessian SDClip, LQER, AWQ~\citep{lin2024awq}, and Crown-Q.

  \item \textbf{Architecture} (37 flags): structural modifications to the transformer or
    non-transformer model. Covers depth recurrence, parallel residuals, XSA, SmearGate, Bigram Hash, U-Net skip, Partial RoPE~\citep{su2021roformer}, YaRN RoPE~\citep{peng2024yarn}, NTK-Aware RoPE, LeakyReLU$^2$, SwiGLU~\citep{shazeer2020glu},
    Gated DeltaNet~\citep{yang2024gateddeltanet}, State-Space Model~\citep{gu2023mamba},
    Universal Transformer, VarLen Attention, Flash Attention~3~\citep{shah2024flashattn3},
    Fused MLP, Fused QKV Projection, Cross-Layer Attention, KV Sharing, AttnOutGate, Paged KV Cache, Gated Attention, Sparse Attention Gate, Logit Softcap, Polynomial Softcap,
    AsymLogit, Mixture of Softmax, MoE MLP, Stochastic Depth, Selective Pruning, Skip Gates,
    Tied Embeddings, Factored Tied Embedding, Value Embeddings, Value Residual, and PPM Byte
    Mixture.

  \item \textbf{Tokenization} (4 flags): vocabulary design and preprocessing choices. Covers
    SP1024, SP4096, SP8192, and CaseOps/Casefold.

  \item \textbf{Training} (16 flags): optimizer and regularization choices. Covers Muon WD~\citep{jordan2024muon}, SWA, EMA, MuonEq-R, Parallel Muon, NeoMuon, OrthoInit, Overtone Init, QK-Gain, Lower LR, Banked Muon, Muon Momentum 0.97, Polar Express NS, Warmdown, Z-Loss, and Complementary Training.

  \item \textbf{Evaluation} (10 flags): strategies applied at inference or evaluation time
    inside the 16\,MB artifact. Covers Sliding Window, the six TTT variants (LoRA, Legal/ScoreFirst, Pre-Quant/Track-A, Global/Phased, SLOT, and Other), N-gram Backoff, N-gram Cache, and Neural Cache.
\end{enumerate}

\paragraph{Techniques referenced in the main text but not described in Section~\ref{sec:phases}.}
\begin{itemize}
  \item \textbf{Attention Output Gate} and \textbf{AsymLogit} are challenge-community techniques: the former applies a learned gate to the attention sublayer's output before it enters the residual stream; the latter is an asymmetric variant of output-logit soft-capping that bounds positive and negative logits differently.
  \item \textbf{Ternary and Binary quantization} restrict weights to $\{-1, 0, +1\}$ or $\{-1, +1\}$, as in BitNet~\citep{ma2024bitnet}.
  \item \textbf{Universal Transformer}~\citep{dehghani2019universal} applies one shared layer recurrently over depth in place of a stack of distinct layers.
  \item \textbf{State-Space Model} replaces attention with a selective linear state-space recurrence, as in Mamba~\citep{gu2023mamba}.
  \item \textbf{Value residuals}~\citep{zhou2024valueresidual} add the first layer's value vectors into every later layer's values, preserving early-layer information.
  \item \textbf{SWA} (stochastic weight averaging)~\citep{izmailov2018swa} averages weights sampled along the training trajectory.
\end{itemize}